\documentclass[11pt,a4paper]{article}
\usepackage{booktabs}
\usepackage{graphicx}

\usepackage[hyperref]{eacl2021}
\usepackage{times}
\usepackage{latexsym}
\usepackage{subfig}
\usepackage{natbib}
\usepackage{notoccite}

\usepackage{microtype}

\aclfinalcopy 

\title{How COVID-19 Is Changing Our Language : Detecting Semantic Shift in Twitter Word Embeddings}

\author{Yanzhu Guo\\
  École Polytechnique \\
\And
  Christos Xypolopoulos\\
  École Polytechnique \\
\And
  Michalis Vazirgiannis\\
  École Polytechnique \& AUEB \\}

\date{}

\begin{document}

\maketitle
\begin{abstract}
Words are malleable objects, influenced by events that are reflected in written texts. Situated in the global outbreak of COVID-19, our research aims at detecting semantic shifts in social media language triggered by the health crisis. With COVID-19 related big data extracted from Twitter, we train separate word embedding models for different time periods after the outbreak. We employ an alignment-based approach to compare these embeddings with a general-purpose Twitter embedding unrelated to COVID-19. We also compare our trained embeddings among them to observe diachronic evolution. Carrying out case studies on a set of words chosen by topic detection, we verify that our alignment approach is valid. Finally, we quantify the size of global semantic shift by a stability measure based on back-and-forth rotational alignment. 

Language is the external manifestation of  human minds. The results of our research provide insight into how the pandemic shapes not only how we write and speak, but also how we think and feel. It is crucial for us to understand these effects in order to deal with them in a more rational and beneficial way.
\end{abstract}

\section{INTRODUCTION}
Words are adaptable to the environment, their senses and meanings are object to constant variations, i.e. semantic shifts. Understanding how they change across different contexts and time periods is crucial in revealing the role of language during social and cultural evolution. 

Semantic shifts can be the consequence of long-term changes of term usage due to political, societal or economic circumstances. An example is the word "gay", whose sense shifted during the 20th century, from meaning "merry" and "cheerful" to eventually "homosexuality". The influence of punctual historic events are equally important and can lead to dramatic changes within a short period of time. For example, the tragic occurrence of the 9/11 attacks has significantly changed the general interpretation of the word "terrorism". 

First detected in Wuhan, China on December 31th 2019, the COVID-19 outbreak grew rapidly in scale and severity, and was officially declared as a pandemic on March 11th, 2020. As one of the most influential global events in modern human history, it is bound to leave its mark on the way we use language as well.

Social media have promoted new forms of social interaction and changed the way people communicate and interact. People use social media to report the latest news, but also to express their opinions and feelings about real-world events. Users show particular interest in emergency situations such as the COVID-19 infection. Each day, users post millions of messages related to the pandemic. Data collected from these platforms can serve as valuable resources while studying the effect of the pandemic on human language. Their diversity and comprehensibility are guaranteed by the inclusive nature of social media. 

In this paper, we explore the semantic stability of words by computing how their meanings, represented by the respective embedding vectors, have been influenced by the pandemic. We first collect COVID-19 related tweets for three consecutive months after the outbreak and used distributional semantics to generate independent embedding spaces for each of these three months. We compare these embedding spaces with a general-purpose Twitter word embedding model generated from tweets collected before the pandemic. We also compare the three COVID-19 related embedding spaces among themselves to detect semantic shift of vectors over time. To ensure that the comparisons are reasonable, we need to align the vectors between different latent spaces. Inspired by \citet{zhang2016past} and \citet{diachronic}, we project words from one embedding space to another, achieving alignment by the creation of an optimal rotational mapping. 

Our main research problem is to study how the COVID-19 pandemic has induced semantic shifts in words, carry out case studies on certain words undergoing significant shifts and quantify the size of global shift.

\section{RELATED WORK}
In this section, we examine related literature falling into two categories : detection of semantic shifts and COVID-19 related Twitter analysis.

\subsection{Detection of semantic shifts}
Vector representations of words generated by distributional methods have become more and more popular ever since the publication of Word2Vec \citet{word2vec}. Recently, many researchers have been applying them in the detection of semantic shifts. Most of the proposed approaches are separated into two steps: first generate separate word embeddings for each context and time period independently, then use mathematical approaches to align the different word embeddings. The quality of the alignment is critical to the results of comparisons. \citet{statistically} computed the optimal linear transformation between the base and target embedding space by solving a least squares problem of k-nearest neighbors. \citet{diachronic} also employed linear transformations for alignment but only considered orthogonal ones. \citet{zhang2016past} achieved alignment through a similar way, adding the usage of anchor words, whose meanings are supposed to remain stable between the two embedding spaces. The choice of anchor words requires prior expert knowledge. In more recent efforts, \cite{temporal} avoided solving a separate alignment problem by learning the word embeddings jointly.

The approach we choose to employ is a combination of \citet{zhang2016past} and \citet{diachronic}. However, we do not determine the anchor words manually. We choose the most frequent words in the general vocabulary as anchor points, relying on "the law of conformity" introduced by \citet{diachronic}.

For evaluating the proposed approach, previous works such as \citet{heyer2009change} and \cite{malleable} selected a set of example words whose shift of meanings are supervised by human. We use a similar approach and select a small set of example words through topic detection by clustering hashtags.

\subsection{COVID-19 related Twitter analysis}
Having tremendously impacted everyone’s daily life, the COVID-19 pandemic has become the center of research interest in 2020. More specifically, numerous papers are focused on how Twitter and other social media users reacted to this health crisis. \citet{chen2020covid} released the first public COVID-19 Twitter dataset. \citet{12} addressed the diffusion of misinformation about COVID-19 on Twitter. \citet{ziems2020racism} reveals the origin and spread pattern of racist online behavior during the outbreak of COVID-19.  \citet{32} established an understanding of people's reactions towards COVID-19 policies by mining a multilanguage Twitter dataset. \citet{bert} released COVID-Twitter-BERT, a pretrained transformer-based model, with a wide range of applications in COVID-19 related NLP tasks.

Regarding semantic shifts, \citet{schild2020go} trained weekly Word2Vec models from Twitter data collected after the outbreak and observed shifts towards appearance of more sinophobic slurs. However, to our knowledge, there has been no systematic or comprehensive study on semantic shifts induced by COVID-19 up to this date.

\section{EXPERIMENTAL SETUP}
\subsection{Datasets}

The tweets used for building our COVID-19 related word embedding models date from April 2020 to June 2020. The objective of this dataset is to collect a huge, coherent body of texts related to the pandemic. In this case, our filters are focused on tweets that include the hashtags “covid19” and “coronavirus”. Through Twitter’s public streaming API, we extracted tweets in English language marked with either or both of the two above hashtags.

Given that the pandemic has been the center of global attention ever since its outbreak and that the situation is evolving rapidly on a daily basis, we have reason to believe that there would be detectable semantic shifts even over the short periods of single months. Therefore, we construct monthly word embeddings for each of the months April, May and June. After removing retweets and quoted tweets, the dataset consists of 35 million unique tweets (2.7GB) for the month of April, 21.0 million (1.7GB) for the month of May and 13.4 million (1.1GB) for the month of May. The total size of vocabulary is 0.3 million. 

We use pre-trained word embeddings twitter-200 from open-source package gensim to serve as a reference semantic space for Twitter language before the outbreak of COVID-19. This embedding is trained on 2 billion tweets dating before 2017, with a vocabulary size of 1.2 million.

\subsection{Preprocessing}
Each tweet used in the training process is preprocessed with the following steps :

\begin{itemize}
    \item \textbf {Conversion to lowercase} : All letters in each tweet is converted to lowercase.

    \item \textbf{Removing URLs and mentions}: We remove all mentions and URLs appearing within the tweets.  
    
    \item \textbf{Removing emoji symbols and special characters}: We remove all emoji symbols and special characters, except hashtag symbols \#.
    
    \item \textbf{Removing stop words}: Stop words are frequent words that carry little information, such as prepositions, pronouns, and conjunctions. We remove words from the list of English stop words in the NLTK library.
    
    \item \textbf{Removing short tweets}: Extremely short tweets, i.e. tweets with fewer then 10 tokens in total, do not contain meaningful content in most cases and are therefore removed.

\end{itemize}

These preprocessing steps are necessary in order to reduce the vector space size and computing cost.

\subsection{Model training}
We use the open source implementation of Word2Vec in the gensim package to generate word embeddings. We remove words that occur less than 10 times and apply the Skipgram architecture. We choose a window size of 5 and dimensionality of 200 for the word vectors.

Within each month, we iterate over 10 epochs. The learning rate is set to 0.025 at the start of each epoch and linearly decreased to 0.0025.

We initialize the word vectors of month April randomly and use tweets from April to train them. Once the training stage has been completed, we use them to initialize the word vectors of May. Similarly, we initialize the word vectors of June with the trained vectors of May.

\section{MEASURING WORD STABILITY}
In order to detect semantic shifts, we need to define a quantitative measurement of each word's stability. We achieve this by first aligning the different word embeddings and then applying two-way rotational mappings.

\subsection{Aligning word embeddings}
A key issue in comparing different word embeddings across different contexts and time periods is alignment. Due to the rotational invariance of cost functions in the Word2Vec training algorithm, the separately-learned embeddings are placed into different latent spaces. This does not affect pairwise cosine similarities within the same embedding space but will obstruct comparison of the same word across context and time. 

Having trained COVID-19 related word embeddings for each of the three months April, May and June, we must now align all these three embeddings with the general-purpose embedding unrelated to COVID-19 so that all of them fit into one unified coordinate space. This enables us to characterize the semantic change between them. 

During the alignment process, we make two simplifying assumptions: First, we assume that the spaces are equivalent under an orthogonal rotation. Second, we assume that the meaning of the most frequent words in the general vocabulary do not shift over time, and therefore, their local structure is preserved. We make the second assumption according to "the law of conformity" proposed by \citet{diachronic}, stating that rates of semantic change scale with a negative power of word frequency.

We use orthogonal Procrustes \citet{schonemann1966generalized} to align the learned embeddings to the same coordinate axes. Specifically, we define $W_0 \in R^{d \times |V|}$ as the matrix of the 1000 most frequent words in the base embedding model. We define $W \in R^{d \times |V|}$ to be the matrix of the same 1000 words in the target embedding model. We align $W$ to $W_0$ while preserving cosine similarities by optimizing:
$$R = arg min_{Q^TQ = I} \|W_0Q-W\|_F$$ 
We solve this optimization problem by an application of singular value decomposition and obtain the best rotational mapping between the two embedding spaces.

\subsection{Two-way Rotational Mappings}

After aligning the embeddings across different contexts and time periods, we use them to measure the stability of words. The less stable a word is, the more semantic displacement it has undergone in the new context.

A natural idea is to directly compute the cosine similarities between a word’s representations in the aligned embedding spaces. However, according to \citet{malleable} and based on our own experiments, the similarity values we obtain after executing only for once the rotational mappings ($sim_{ij}(w) = cos-sim(R^{ij}V^i_w,V^j_w)$ ) are low on average, indicating that one-way rotational mappings are not of high quality. 

Nevertheless, when we use two-way rotational mappings, the average similarity increases significantly. Words are generally mapped close to their starting orientations when mapped back into their original space. Inspired by \citet{malleable}, we define the ‘stability’ of a a word $w$ as the following measure:
$$stab(w) = \frac{sim_{01}(w) + sim_{10}(w)}{2}$$
$$sim_{ij}(w) = cos-sim(R^{ji}R^{ij}V^i_w,V^i_w)$$
The stability of a word using this measure equals to the similarity of its vector to its mapped vector after applying the rotational mapping back and forth.

\section{TOPIC DETECTION AND CASE STUDIES}

We perform topic detection among the collected COVID-19 related tweets as words highly involved in relevant topics are more likely to endure significant semantic shifts after the outbreak. We then carry out case studies on words from the detected topics.

\subsection{Clustering hashtags}
\begin{figure*}[ht]
	\centering
	\includegraphics[width=0.85\textwidth]{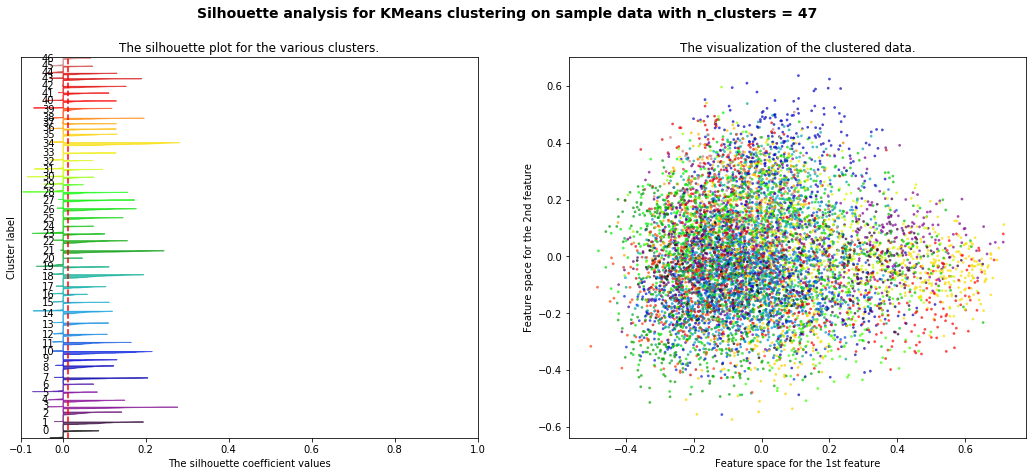}
	\caption{Silhouette coefficients and PCA decomposition of optimal clustering.}
	\label{fig2}
\end{figure*}

Hashtag is a type of metadata tag used on social networks such as Twitter. It lets users apply dynamic, user-generated tagging that helps other users easily find messages with a specific theme or content. We do not remove hashtags from tweets in the preprocessing step. We treat them in the exact same way as other words from the vocabulary during training. Thus, they are positioned into the embedding vector space such that hashtags sharing common contexts in the corpus are located close to one another. We normalize the hashtag vectors and perform K-means algorithm, in the hope of observing distinctive topics. Using silhouette analysis \citet{rousseeuw1987silhouettes}, we determine the optimal number of clusters to be 47. We also implement PCA to transform the computed 200-dimensional word vectors into 2-dimensional points in order to visualize the clusters. The silhouette plot and PCA decomposition is shown in Figure \ref{fig2}.

Silhouette analysis is used to study the separation distance between the resulting clusters. The silhouette plot displays the silhouette coefficients of each resulting cluster which are measures within the range of $[-1, 1]$. Higher values indicate that the samples are far away from the neighboring clusters and that the  resulting clusters are coherent within themselves. 

We do not observe a clear separation in 2-dimensional space but the silhouette plot shows the clusters are coherent in general. In addition, analyzing the top hashtags in each cluster gives distinctive topics. 

\begin{figure}[ht]
	\centering
	\includegraphics[width=0.45\textwidth]{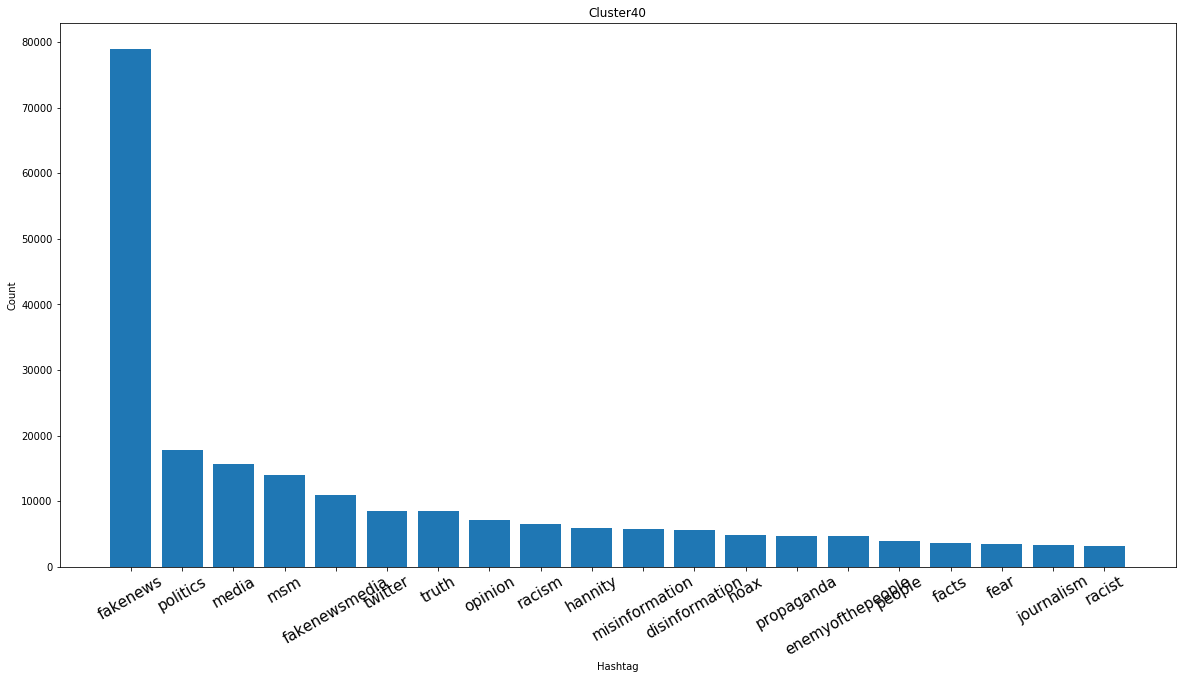}
	\caption{Topic : Racism.}
	\label{fig21}
\end{figure}

\begin{figure}[ht]
	\centering
	\includegraphics[width=0.45\textwidth]{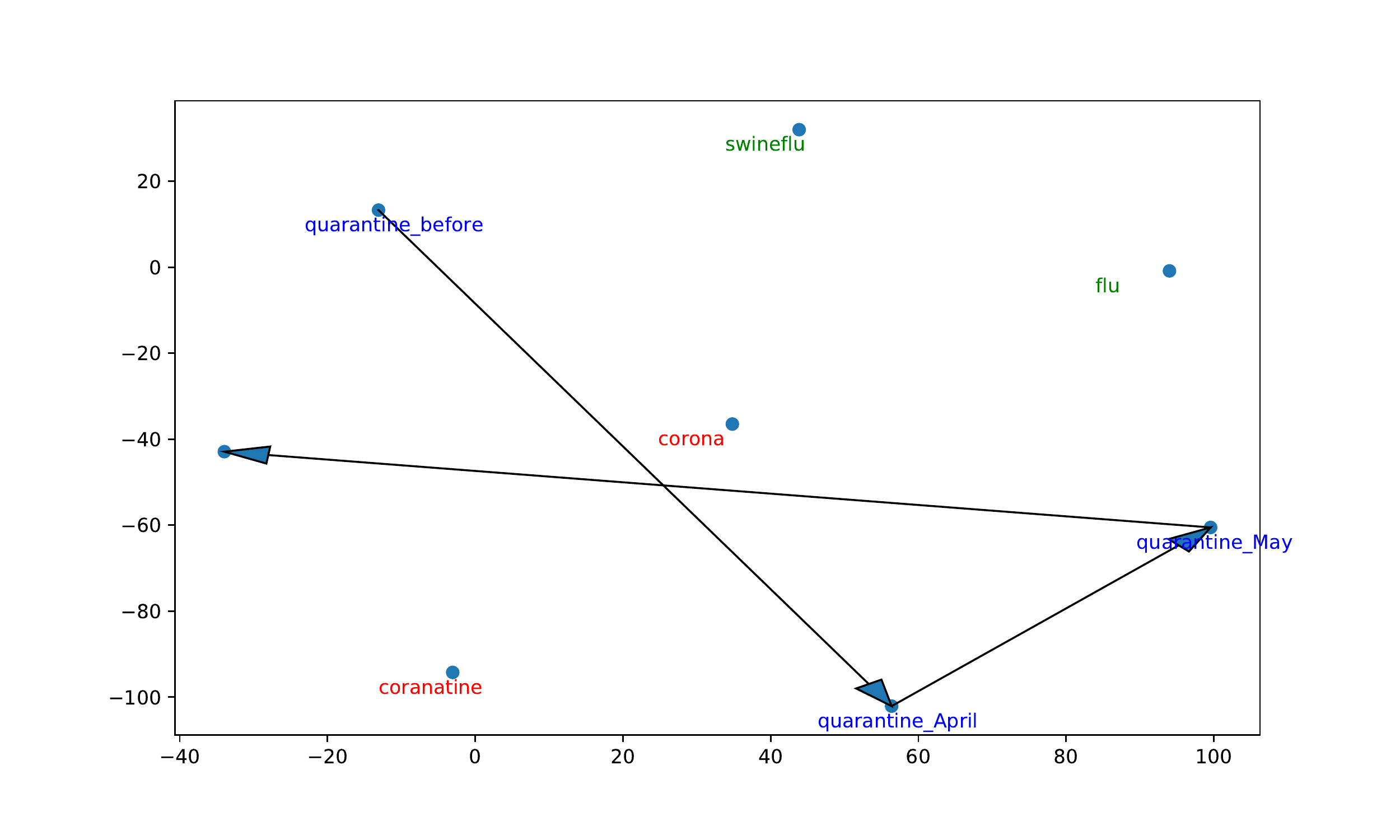}
	\caption{Topic : Quarantine.}
	\label{fig22}
\end{figure}

\begin{figure}[ht]
	\centering
	\includegraphics[width=0.45\textwidth]{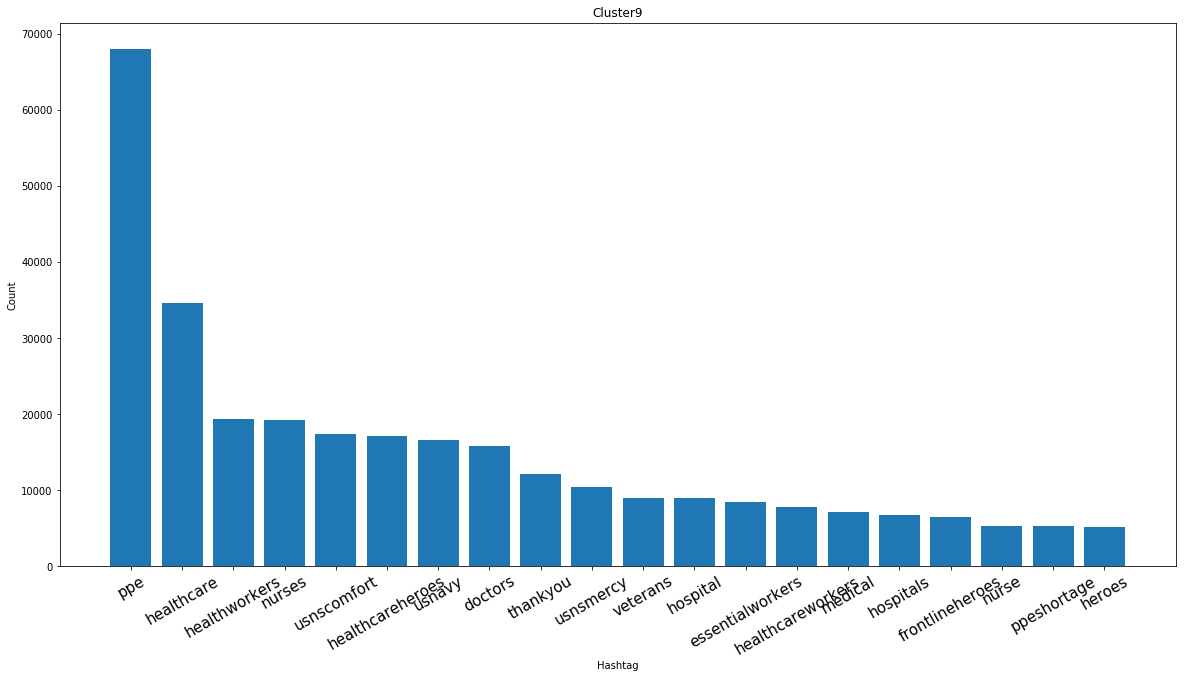}
	\caption{Topic : Health workers.}
	\label{fig23}
\end{figure}

\begin{figure}[ht]
	\centering
	\includegraphics[width=0.45\textwidth]{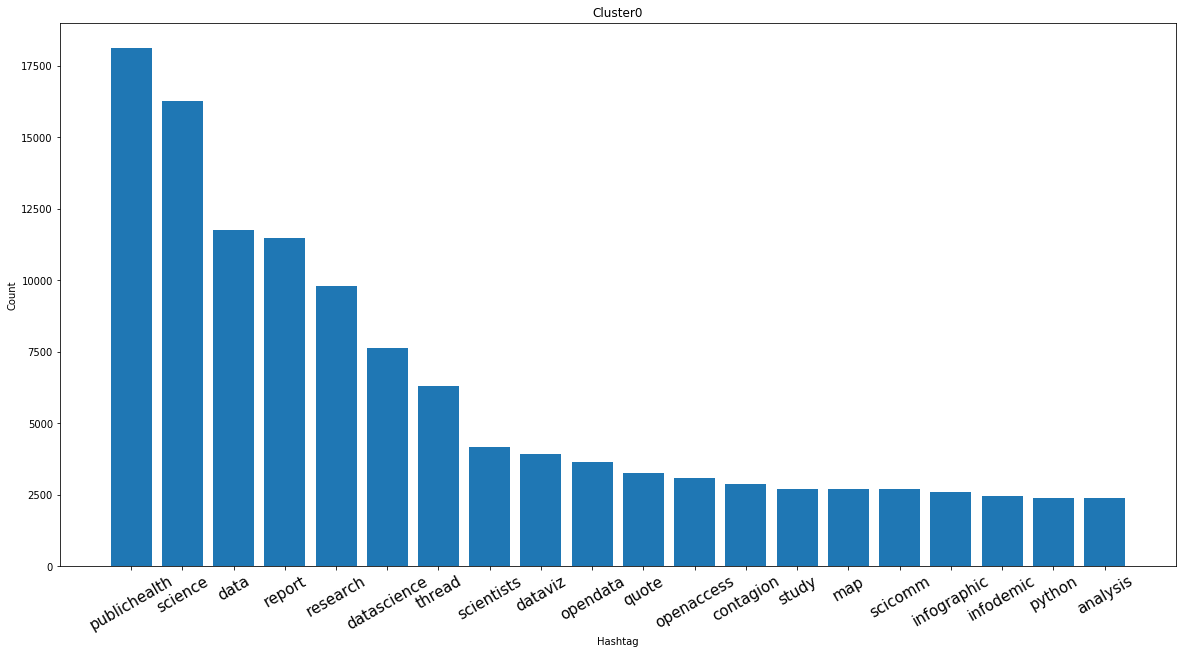}
	\caption{Topic : Artificial intelligence.}
	\label{fig24}
\end{figure}

 In Figures \ref{fig21} to \ref{fig24}, we show 4 examples of topics on which we will carry out case studies in the following. The contents of the top hashtags are plotted along the x axis while the y axis represents the number of times a hashtag appears in the whole dataset.
 
 The cluster shown in Figure \ref{fig21} is centered around racism. Fake news and racist comments have been circulating wildly on social media. After investigating their patterns, we find most of them to be closely associated with China and Asia.
 
 The cluster shown in Figure \ref{fig22} is mostly about quarantine. It’s obvious why quarantine appears as a topic in COVID-19 related tweets. However, it’s interesting that we find hashtags such as \#MondayMotivation, \#WednesdayWisdom and \#ThursdayThoughts in this cluster. These are quotes that people use to share positive energy on each day of the week. It seems that people are especially in need to share positive energy during the quarantine period.

The cluster around health workers shown in Figure \ref{fig23} is mostly made up of positive hashtags such as \#ThankYou and \#Heroes. However, the top hashtag is \#ppe, indicating the shortage of personal protection equipment for health workers.

The last cluster shown in Figure \ref{fig24} is dedicated to artificial intelligence and data science. In fact, AI and data science have played important roles in the battle against COVID-19, especially in terms of ehealth.

\subsection{Case Studies}

\begin{figure*}[ht]
	\centering
	\subfloat[racism]{
	\includegraphics[width=0.5\textwidth]{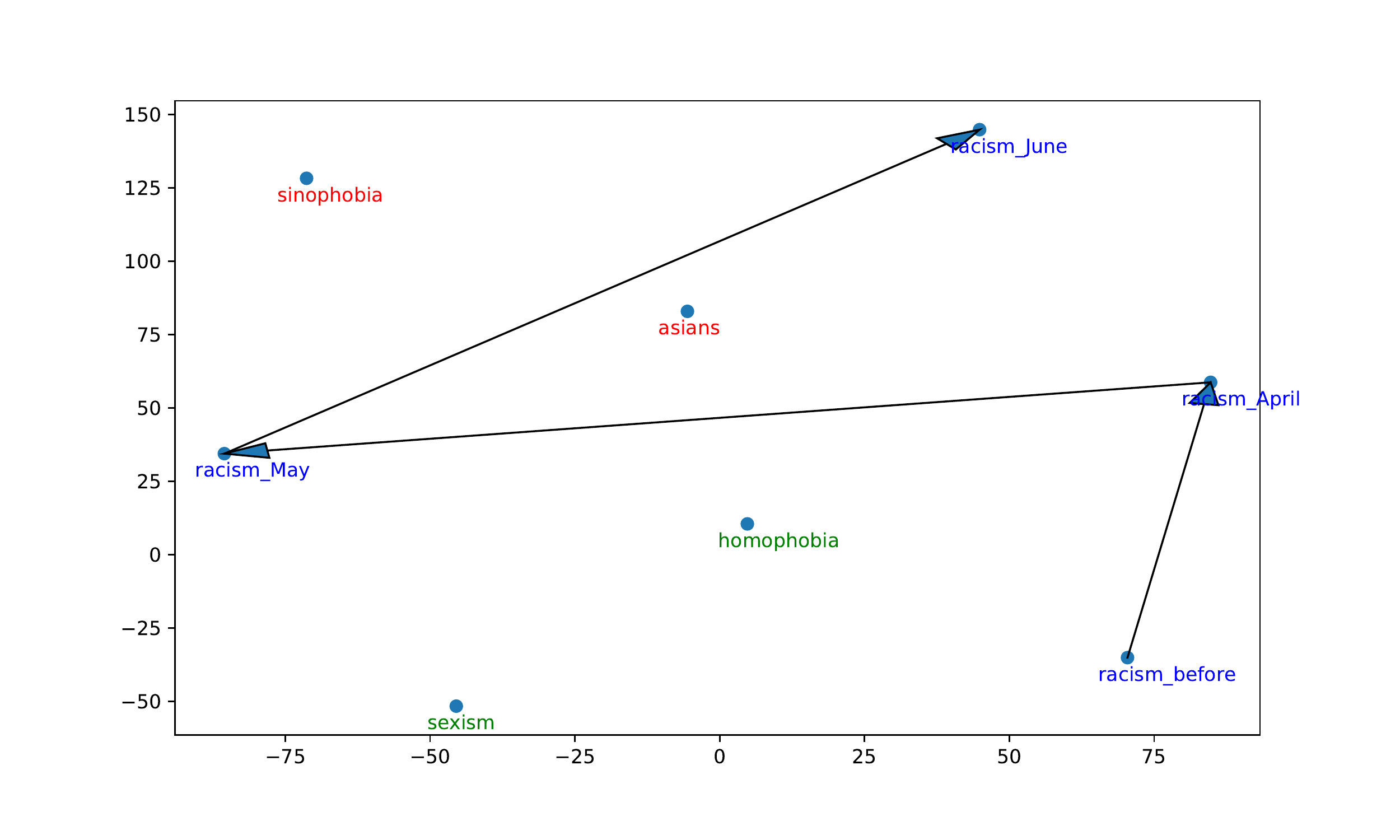}
	}
	\subfloat[quarantine]{
	\includegraphics[width=0.5\textwidth]{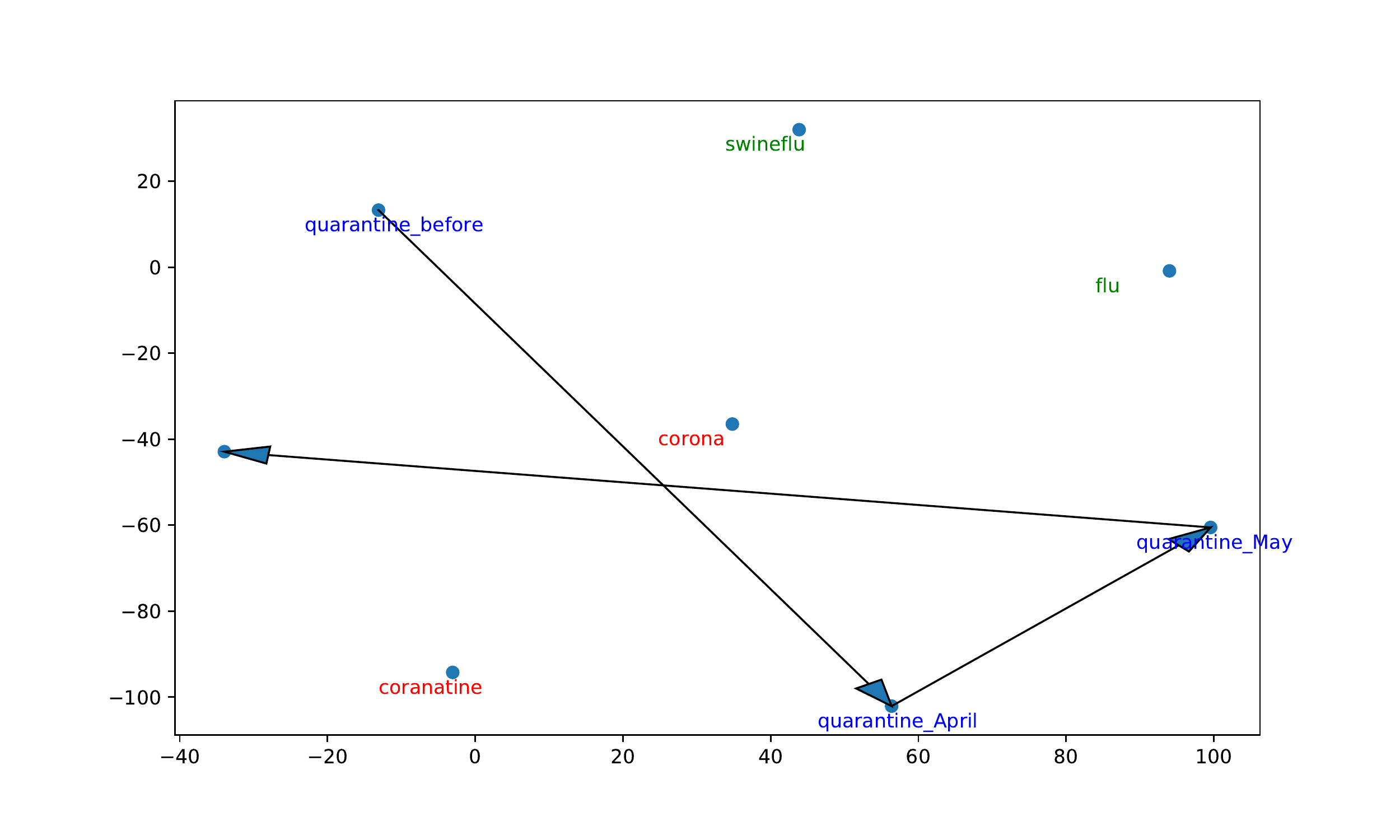}
	}\\
	\subfloat[hero]{
	\includegraphics[width=0.5\textwidth]{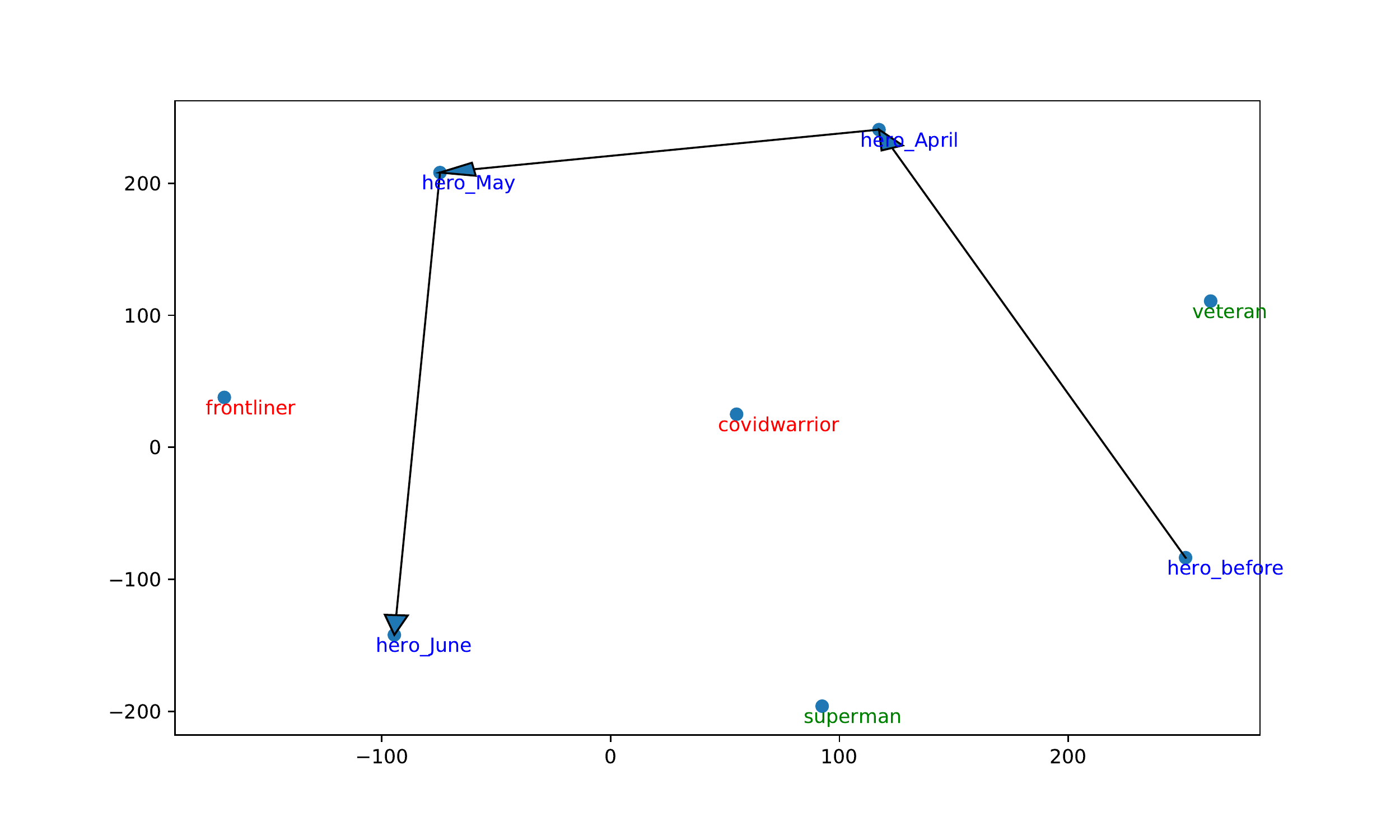}
	}
	\subfloat[ai]{
	\includegraphics[width=0.5\textwidth]{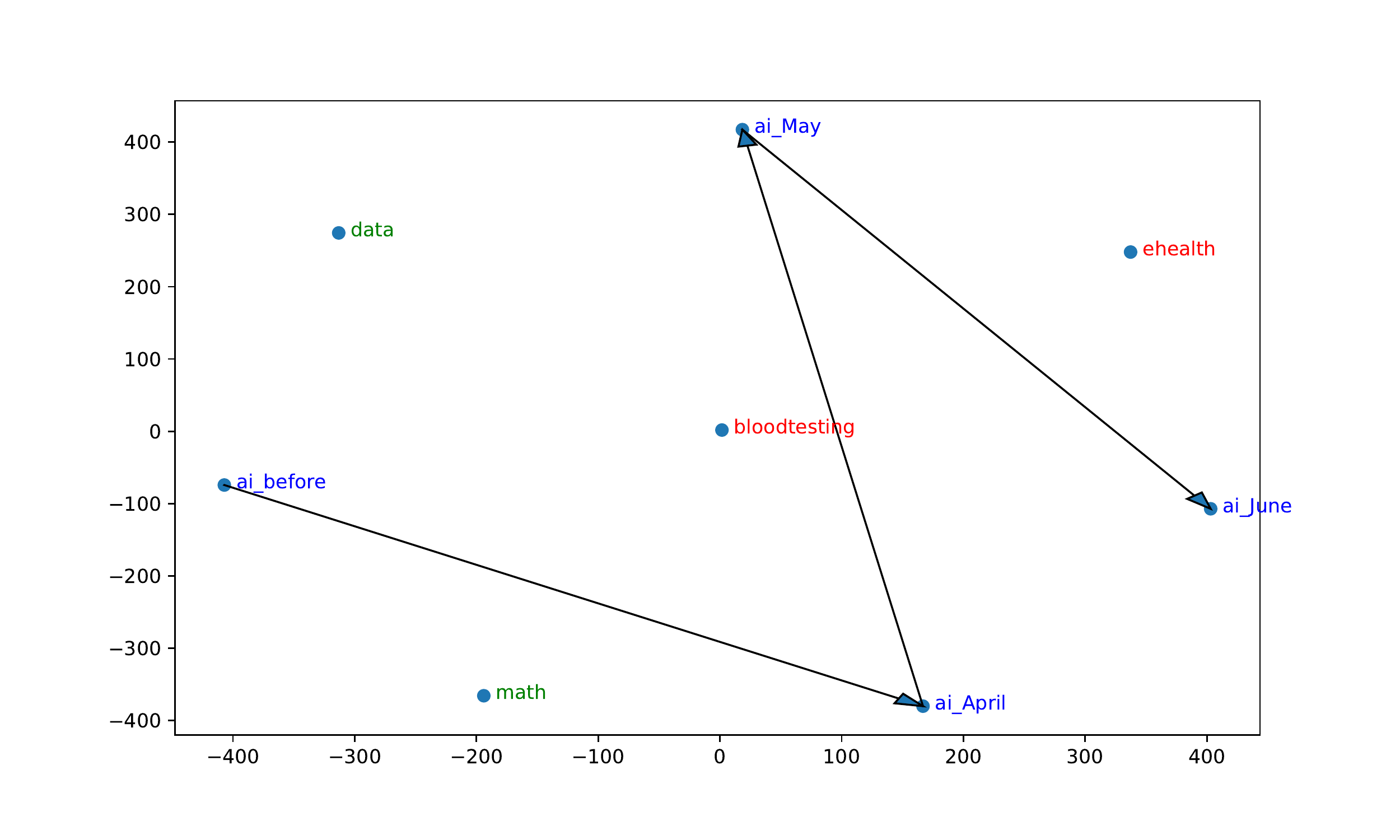}
	}
	\caption{t-SNE visualization of semantic shifts.}
	\label{figure1}
\end{figure*}

\begin{table*}[ht]
\centering
\begin{tabular}{@{}l|ll@{}}
\toprule
Word       & Moving away        & Moving towards           \\ \midrule
racism     & sexism, homophobia & asians, sinophobia       \\
hero       & veteran, superman  & frontliner, covidwarrior \\
quarantine & swineflu, flu      & coranatine, corona       \\
ai         & math, data         & ehealth, bloodtesting    \\ \bottomrule
\end{tabular}
\caption{Set of words with detected semantic shifts}
\label{table1}
\end{table*}

\begin{figure*}[ht]
	\centering
	\subfloat[racism]{
	\includegraphics[width=0.29\textwidth]{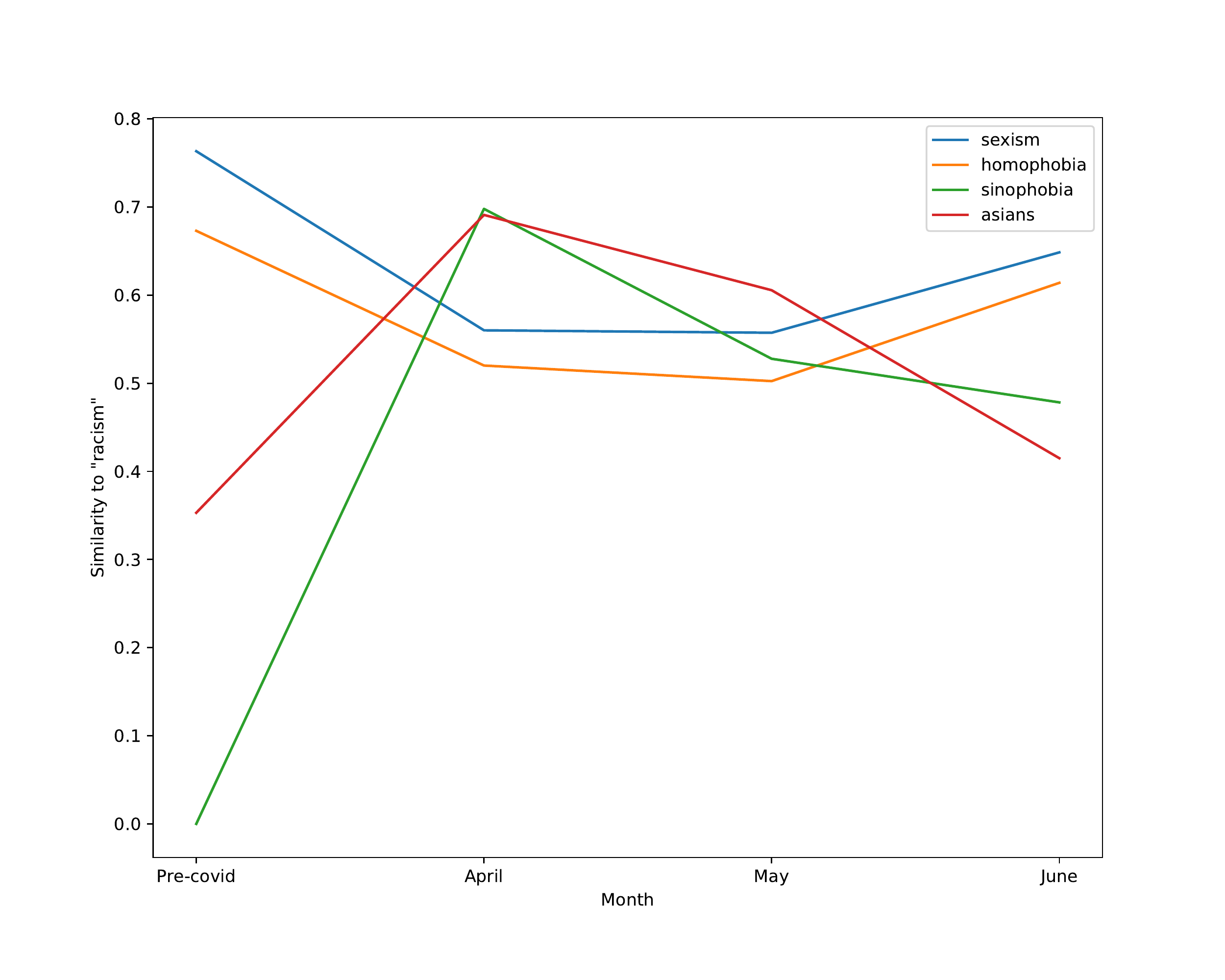}
	}
	\subfloat[quarantine]{
	\includegraphics[width=0.29\textwidth]{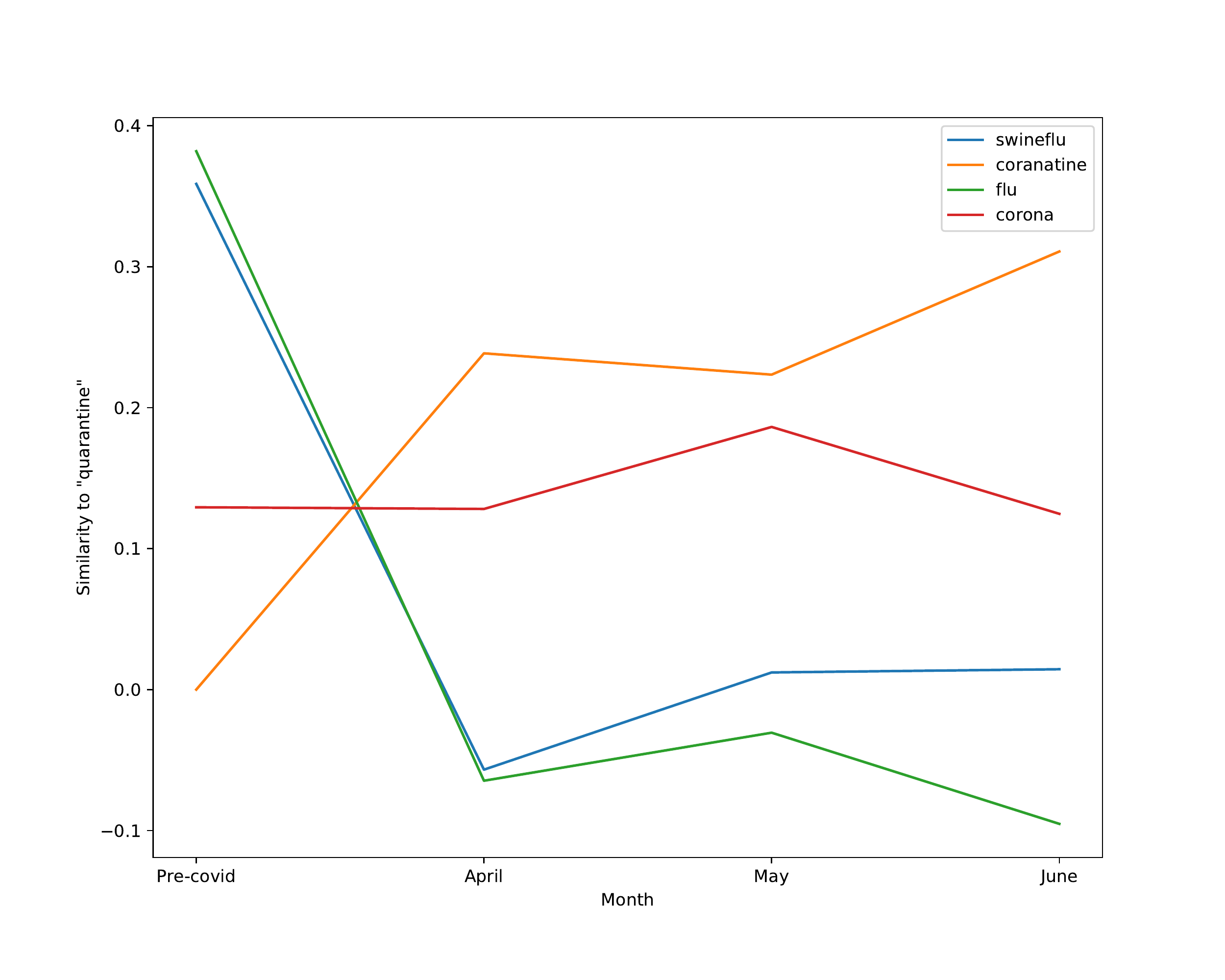}
	}\\
	\subfloat[hero]{
	\includegraphics[width=0.29\textwidth]{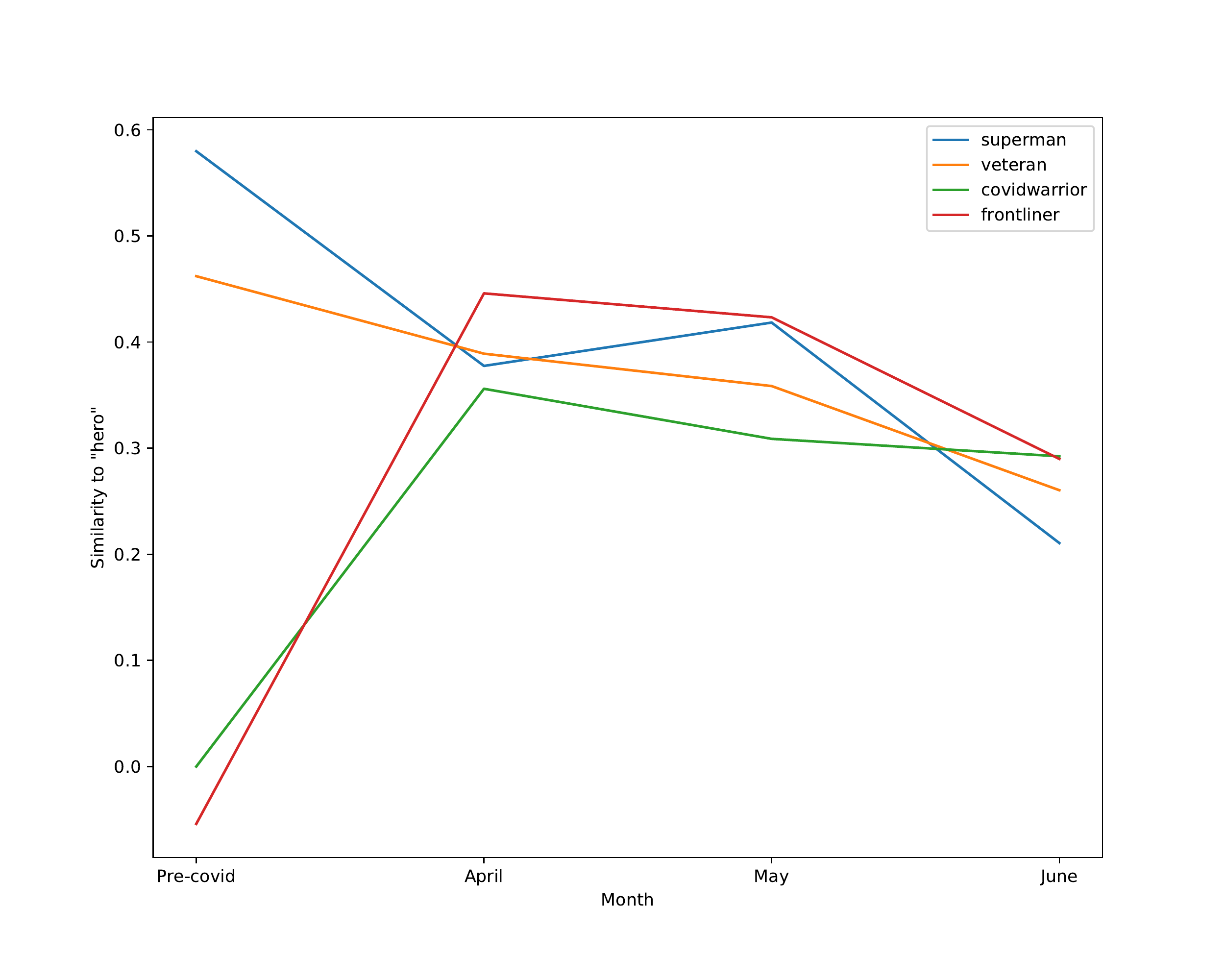}
	}
	\subfloat[ai]{
	\includegraphics[width=0.29\textwidth]{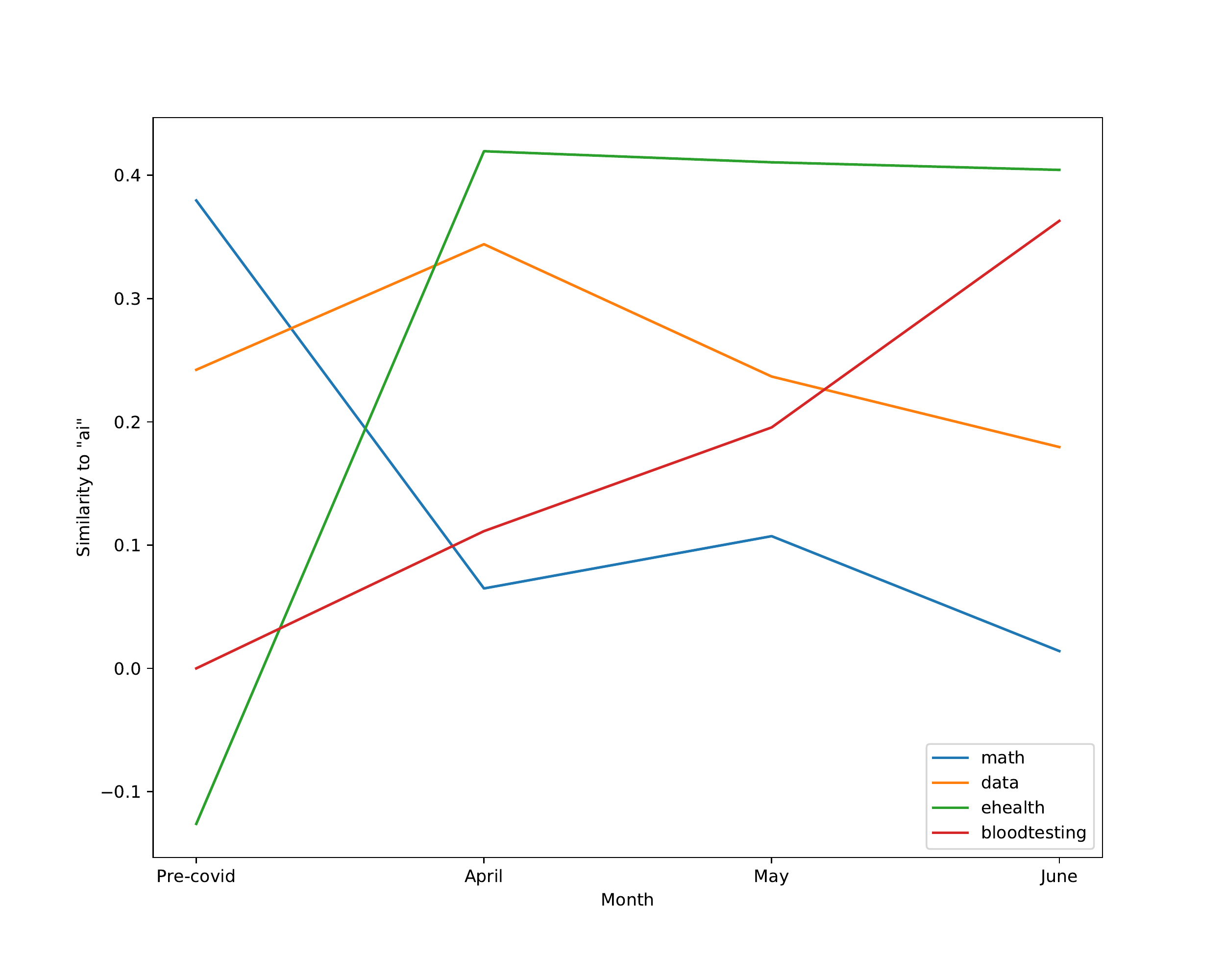}
	}
	\caption{Similarity changes between semantically-shifted words and their surroundings.}
	\label{figure2}
\end{figure*}

Visualizing the trajectory of words in the aligned embedding spaces helps us understand the semantic shifts of words across contexts and over time. Table \ref{table1} summarizes the semantic shifts of a set of example words while figure \ref{figure1} shows their trajectories. We plot the 2-dimensional t-SNE projection of the target word's embedding in each of the aligned models. We also plot some of the surrounding words of the target word in aligned models. Figure \ref{figure2} illustrates the cosine similarity changes between the word of interest and surrounding words. Words that do not exist in the vocabulary of a certain model are assigned a similarity of $0$ to all the other words within this given model.

We pick four words of interest: racism, quarantine, hero and ai. In all cases, the trajectory illustrations and similarity measures demonstrate that the words of interest have shifted significantly in meanings after the outbreak of COVID-19 and also show evolution over the three months. We see "racism" move away from other general concepts of discrimination and end up close to words explicitly expressing hatred towards the Chinese/Asian community. This coincides with the worldwide anti-Asian phenomenon observable since the very beginning of the pandemic. Interestingly, we see in figure \ref{figure2} that the similarity between racism and anti-Asian concepts spiked in April and started decreasing mildly in May and June, indicating that people are slowly gaining back their rationality as the stage of COVID-19 progresses.

\section{RESULTS}

We discuss our results from two perspectives : the stability distributions which indicate the existence of semantic shifts, and the correlations between word frequency and stability which prove the validity of statistical laws.

\subsection{Stability Distributions}

\begin{figure}[ht]
	\centering
	\includegraphics[width=0.45\textwidth]{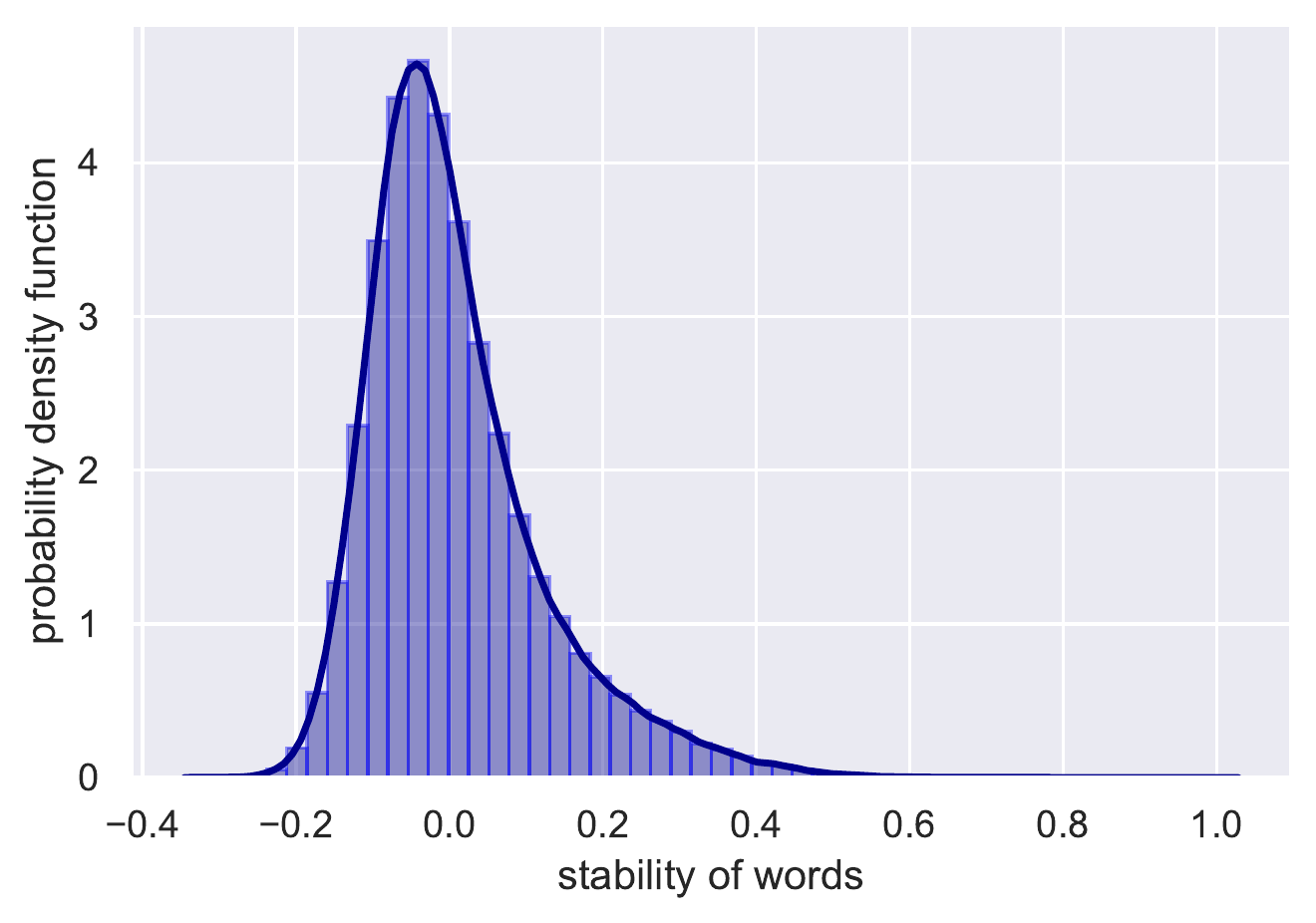}
	\caption{Cross-corpora comparison: stability distribution before and after COVID-19.}
	\label{fig8}
\end{figure}

\begin{figure}[ht]
	\centering
	\includegraphics[width=0.45\textwidth]{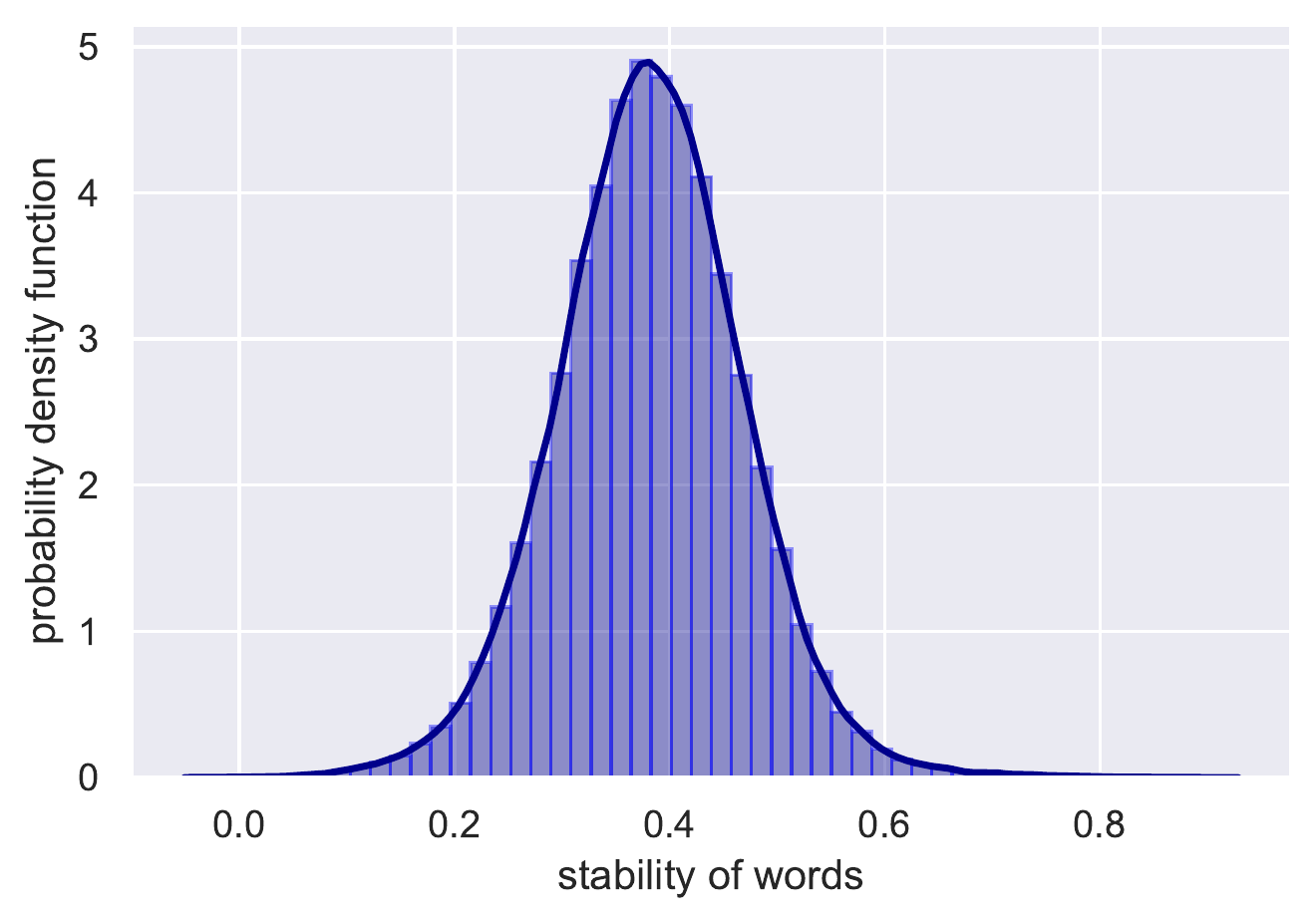}
	\caption{Diachronic evolution: stability distribution during April, May and June.}
	\label{fig9}
\end{figure}

Figure \ref{fig8} shows the stability distribution of words before and after COVID-19. For every word, we compute its stability between the pre-COVID-19 model and each of the three post-COVID-19 models, taking the average value as its final stability measure. We see that the distribution is skewed, centered around 0 and tilted towards the right. 

Figure \ref{fig9} shows the stability distribution of words during the months April, May and June. For every word, we compute separately its stability between April-May and May-June, taking the average value as its final stability measure. We see that the distribution is almost normal, centered around 0.4 and not tilted.

The absolute value of the stability measure does not make sense, considering that the alignment cannot reach perfection even after optimization. Nonetheless, the shapes of the distributions indicate that there exists both cross-corpora and diachronic semantic shifts. Otherwise, the distributions should be extremely left-skewed and concentrated around a relatively high value. Comparing the two distributions, we can reach the conclusion that the semantic change across corpora is more significant than that over monthly time periods. 

\subsection {Statistical laws of semantic change}

Recently, \citet{diachronic} proposed two quantitative laws of semantic change: 
\newline
1) the law of Conformity which implies that “the rate of semantic change scales with an inverse power-law of word frequency”. 
\newline
2) the law of Innovation which reflects that “the semantic change rate of words is highly correlated with their polysemy”. 

\begin{figure}[ht!]
	\centering
	\includegraphics[width=0.45\textwidth]{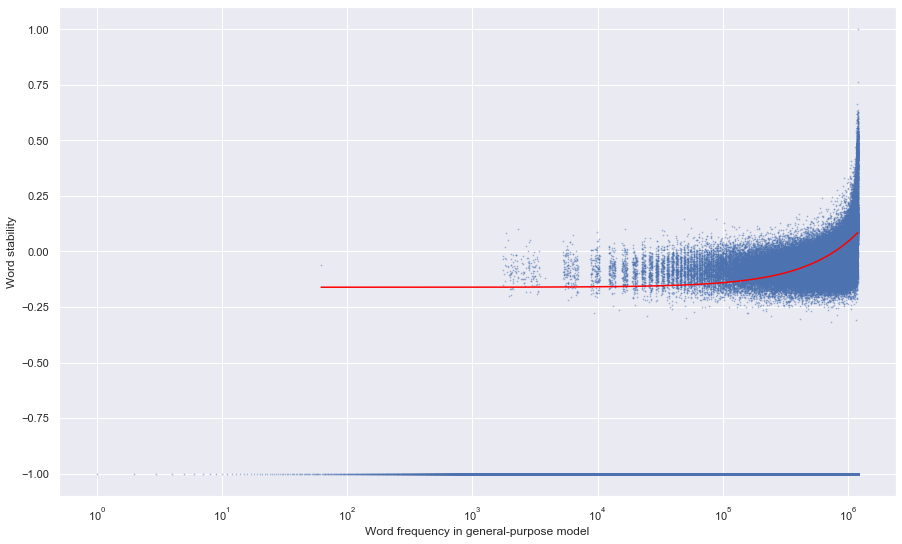}
	\caption{Correlation between word stability and pre-COVID-19 word frequency.}
	\label{fig10}
\end{figure}

\begin{figure}[ht!]
    \centering
	\includegraphics[width=0.45\textwidth]{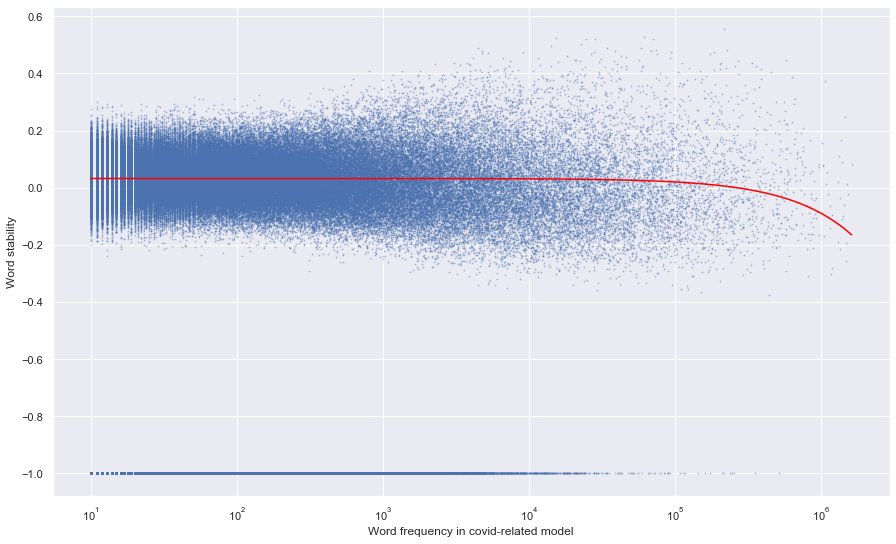}
	\caption{Correlation between word stability and post-COVID-19 word frequency.}
	\label{fig11}
\end{figure}

The stability measure in both Figure \ref{fig10} and Figure \ref{fig11} refers to the cross-corpora stability. The line with value $-1$ corresponds to words that do not exist in at least one of the compared models.

Figure \ref{fig10} plots the correlation between a word's stability and its number of occurrence in the pre-COVID-19 general-purpose Twitter dataset. It proves that "the law of conformity" holds in our case. This may be explained as the result of our choice of anchor words : we use highly frequent words as static points for the computation of mappings. However, we only use the top 1000 frequent words while "the law of conformity" is valid for far more than 1000 data points, as shown in Figure \ref{fig10}. 

Figure \ref{fig11} plots the correlation between a word's stability and its number of occurrence in the COVID-19 related Twitter dataset. Contrary to "the law of conformity", stability is negatively correlated with word frequency here. This reveals a new law of semantic change which is natural to our instincts: words occurring more frequently in the event-specific dataset are more likely to shift meanings.

\section{CONCLUSIONS}

In this work we performed a comparative semantic analysis on four different word embedding models trained before or during the COVID-19 global pandemic. Such a timeframe is a good benchmark for studying semantic changes induced by emergency events in a crisis-ridden society.

Our main contributions are: 
\newline
(1) We showed that the COVID-19 pandemic has introduced noticeable semantic changes in Twitter language and that the fluctuations are continuous from April to June.
\newline
(2) We advanced the rotational mapping approach proposed by \citet{statistically} for detecting semantic shifts, relying on "the law of conformity" for semantic changes discovered by \citet{diachronic}. 
\newline
(3) We further demonstrated that "the law of conformity" holds for semantic shifts induced by specific historical events. On top of that, we introduced a new law of semantic change : words occurring more frequently in the event-specific dataset are more likely to shift meanings.
\newline
(4) We constructed a COVID-19 related Twitter dataset and trained three sets of word embeddings across monthly time periods.

\bibliography{references}
\bibliographystyle{acl_natbib}

\end{document}